  \documentclass{gretsi}

   \usepackage[english,french]{babel}   
  \usepackage{times}			
 
\usepackage{graphicx}
\usepackage{amsmath}
\usepackage{amssymb}
\usepackage{booktabs}
\usepackage{amsthm}
\usepackage{mathtools, cuted}
\usepackage[table,xcdraw]{xcolor}

\usepackage[pagebackref,breaklinks,colorlinks]{hyperref}

\newcommand{\compose}{\circ}

\newtheorem{lemma}{Lemma}

\usepackage{algorithm} 
\usepackage{algpseudocode}
\usepackage{booktabs}
\usepackage{tabularx}
\usepackage{array}
\usepackage{bm}
\usepackage{url}
\usepackage{bbm}
\usepackage{enumitem}
\usepackage{stackengine}
\stackMath
\usepackage{lipsum}

\usepackage[capitalize]{cleveref}
\crefname{section}{Sec.}{Secs.}
\Crefname{section}{Section}{Sections}
\Crefname{table}{Table}{Tables}
\crefname{table}{Tab.}{Tabs.}
 
\title{Defining an action of $SO(d)$-rotations on images generated by projections of $d$-dimensional objects:\\

Applications to pose inference with Geometric VAEs}

\author{\coord{Nicolas}{Legendre}{1,2},
        \coord{Khanh}{Dao Duc}{2}
        \coord{Nina}{Miolane}{3},
    }

\address{\affil{1}{Department of Mathematics, Centrale Supelec\\3 Rue Joliot Curie, Gif-sur-Yvette, 91190, France}
\affil{2}{Department of Mathematics, University of British Columbia\\ 1984 Mathematics Road, Vancouver, BC V6T 1Z4, Canada}
         \affil{3}{Department of Electrical and Computer Engineering \\
         Harold Frank Hall, Santa Barbara, California, 93106, United States}
         }

\email{nicolas.legendre@student-cs.fr, kdd@math.ubc.ca (corresponding author), ninamiolane@ucsb.edu (corresponding author)\\
}

\frenchabstract{Les récents progrès dans le domaine des autoencodeurs variationnels (VAEs) ont permis l'apprentissage de variétés latentes sur des groupes de Lie compacts, tels que $SO(d)$. Une telle approche supposant l'espace des données homéomorphe au groupe de Lie, nous étudions ici la validité de cette hypothèse dans le contexte d'images générées par projection d'un volume de dimension $d$, dont la pose dans $SO(d)$ est inconnue. Après examen de différents candidats définissant l'espace des images et groupe, on montre que l'on ne peut de manière générale obtenir une action de groupe, sans une contrainte  supplémentaire sur le volume. En appliquant des VAEs géométriques, nos expériences confirment que ces contraintes géométriques sont essentielles pour l'inférence de la pose associée au volume projeté, et nous discutons pour conclure des applications potentielles de ces résultats.}

\englishabstract{Recent advances in variational autoencoders (VAEs) have enabled learning latent manifolds as compact Lie groups, such as $SO(d)$. Since this approach assumes that data lies on a subspace that is homeomorphic to the Lie group itself, we here investigate how this assumption holds in the context of images that are generated by projecting a $d$-dimensional volume with unknown pose in $SO(d)$. Upon examining different theoretical candidates for the group and image space, we show that the attempt to define a group action on the data space generally fails, as it requires more specific geometric constraints on the volume. Using geometric VAEs, our experiments confirm that this constraint is key to proper pose inference, and we discuss the potential of these results for applications and future work. }

\begin{document}
\maketitle

\section{Introduction}

 Variational Autoencoders (VAEs) are deep generative models that have been successfully applied across fields to infer latent variables associated with raw data  \cite{Kingma2014Auto-EncodingBayes}. While the traditional VAE architecture was introduced for latent spaces that are homeomorphic to $\mathbb{R}^L$, more recent developments have extended this architecture to latent spaces homeomorphic to Lie groups, with proper reparametrization trick and decoder that respect the group structure \cite{falorsi2018explorations}. These geometric VAEs hold promises for problems where the data is generated from Lie groups, such as the special orthogonal group of rotations in $d$ dimensions $SO(d)$. For example, in the context of structural biology and cryogenic electron microscopy (cryo-EM), 2D images of biomolecules get collected through a generative process that involves the action of the 3D rotations $SO(3)$ (or poses) on a volume. The overarching goal of cryo-EM studies is to reconstruct this 3D volume from a set of 2D images with unknown pose \cite{MiolanePoitevin2020CVPR,donnat2022deep}. While geometric VAEs offer a natural framework for inferring the pose parameters as elements of $SO(3)$, and accurately reconstructing the 3D volume, the method proposed by Falorsi \emph{et al.} also relies on the key assumption that the data lies on a subspace that is homeomorphic to the Lie group $SO(3)$ itself \cite{falorsi2018explorations}.
 
In this paper, we focus on investigating if this key assumption holds, when the data is generated by the action of $SO(d)$ on a volume, followed by its projection along a fixed axis --- a model akin to the cryo-EM setting. We introduce the image formation model, and find that upon considering various candidates to define the image space, the projection generally prevents a group action from being well defined, requiring some geometric constraints on the volume. We specify these constraints and show a practical construction for such volumes in $SO(d)$. Using geometric VAEs, our experiments confirm that these constraints are key to perform proper pose inference. We discuss the potential of these results for applications and future work. 

\section{Group action on image space}

\subsection{Oriented volume and image formation}

We introduce the mathematical background related to the reconstruction of volumes from their projections. Given a dimension $d >0$, we consider a compact domain $\Omega^{d} \subset \mathbb{R}^{d}$, and define a \emph{volume of reference} $V$ as an element of $\mathcal{V}$, the set of positive distributions on $\Omega^{d}$. For a rotation given by $R\in SO(d)$, the \emph{oriented volume} $R\cdot V$, is a positive distribution on $\Omega^{d}$ such that for all $\mathbf{x}=(x_1,\ldots,x_d)\in \Omega^{d}$
\begin{equation}\label{eq:rotation}
    R \cdot V(\mathbf{x}) = V(R^{-1}(\mathbf{x})).
\end{equation}
The orientation $R$ defines the \textit{pose} of the oriented volume.

We define the \emph{image} associated with the oriented volume $R\cdot V$ as the projection $P_{x_d}$ of the oriented volume on the hyperplane ($x_d=0$), given for all $ (x,\ldots,x_{d-1}) \in \Omega^{d-1}$ by:
\begin{equation}\label{eq:image_formation}
     \quad P_{x_d}\left[ R \cdot V \right](x,\ldots,x_{d-1}) =\int_\Omega R \cdot V (\mathbf{x})dx_d.
\end{equation}
$P_{x_d}\left[ R \cdot V \right]$ is the $d$-dimensional Radon transform of the oriented volume \cite{natterer2001mathematics}, notably used for $d=2$ in CAT scan and computerized tomography \cite{natterer2001mathematics}, and for $d=3$ in cryo-EM where $V$ is a 3D biomolecule.

 In what follows, we consider a volume of dimension $d\geq 2$, and explore if one can define a faithful action of a compact Lie group such as $SO(d)$ on the space of projected images. 
Such an action indeed allows us to show that the space of generated images is homeomorphic to $SO(d)$ and hence apply the framework of geometric VAEs \cite{falorsi2018explorations}.

\subsection{Failures to define a group action}

We first show how two natural attempts to define a (left) group action of $SO(d)$ on images generated by the model in Eq.~\eqref{eq:image_formation} fail due to the projection operator $P$. Generally, any $SO(d)$-action on an image space $M$ is defined by a map $\rho: SO(d) \times M  \rightarrow M$ written as $\rho(R,I) = R \cdot I $
that verifies the defining axioms: 

$\bullet$ Identity: $\forall I \in M,$ and $Id$ the identity element of $SO(d)$:
$$Id \cdot I =I.$$

$\bullet$ Compatibility: $\forall R_1,R_2 \in SO(d) ^2$, $ \forall I \in M$, with $\circ$ denoting the group law of $SO(d)$.
$$R_1 \cdot(R_2 \cdot I)= (R_1\compose R_2) \cdot I.$$

\begin{figure}[h!]
\centering
\includegraphics[scale=0.45]{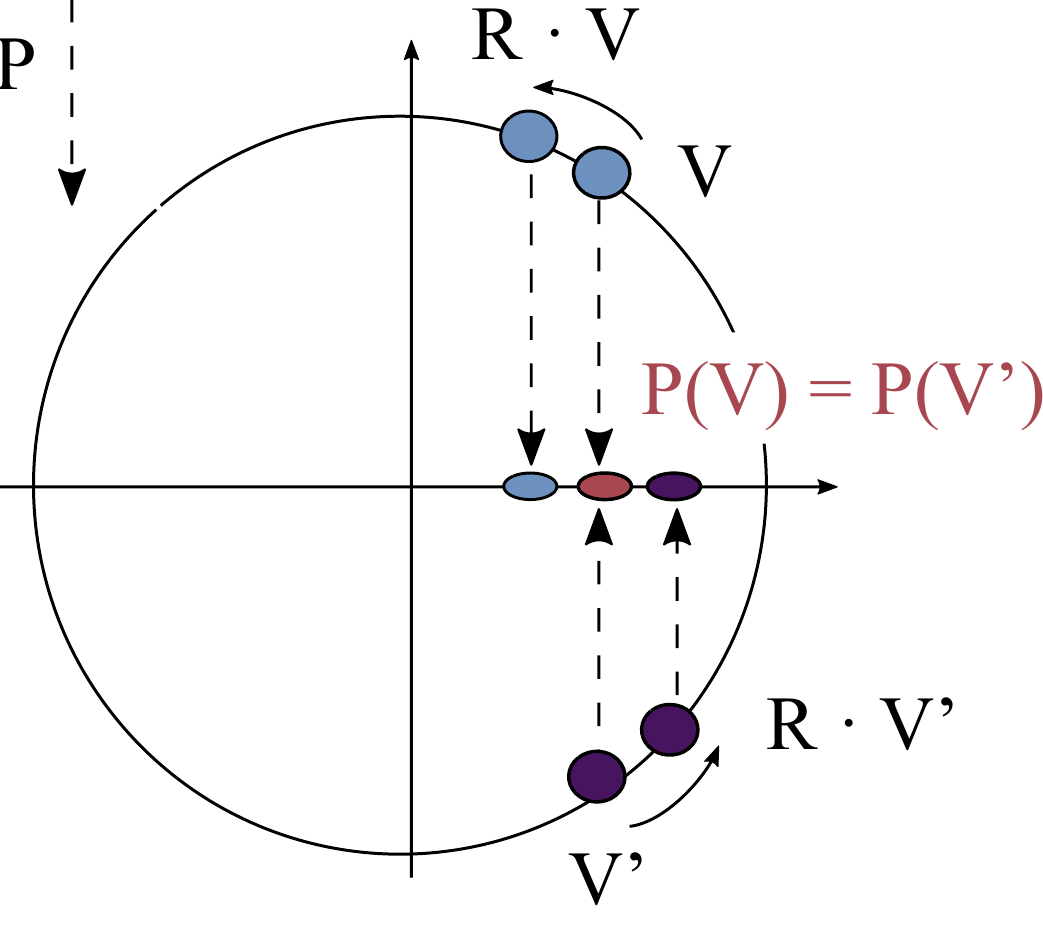}
    \caption{Schematic representation of the main obstacle upon defining a group action on the space of cryo-EM images. For 2 elements $V$ and $V'$ representing two different orientations (represented here as $SO(2)$) of a same volume with identical projections (with the projection operator $P$ represented by a dashed arrow), the action of an additional rotation $R$ is ill-defined on  $I = P[R \cdot V] = P[R \cdot V']$, as it is multi-valued.}
    \label{fig:proj}
\end{figure}

First, we note that $SO(d)$ naturally defines a group action on the space $M=\mathcal{V}$ of $d$-dimensional volumes via their rotations defined in Eq.~\eqref{eq:rotation}. Yet, $SO(d)$ does not define a group action on images seen as singular distributions on $\Omega^d$ and thus as elements of $M=\mathcal{V}$. The projection in Eq.~\eqref{eq:image_formation} makes the identity axiom fail, as the projection of the volume is different from the volume itself. Second, we can attempt to define a $SO(d)$ group action on the space $M=\mathcal{I}_V$ of images defined by Eq.~\eqref{eq:image_formation} for a given volume $V$ as:

\begin{align*}
    \mathcal{I}_V &= \left\{ 
        I \in L_2(\Omega^{(d-1)}) | \exists R_I \in SO(d) , 
            I = P[R_I \cdot V] \right\},  
\end{align*}
through the map: $\rho: SO(d) \times \mathcal{I}_V \rightarrow \mathcal{I}_V$ given by $\rho(R, I) = P((R\circ R_I) \cdot V)$ with $R_I$ one rotation provided thanks to the definition of $\mathcal{I}_V$. In this attempt, the projection leads to the issue illustrated in Figure~\ref{fig:proj}. If we have two rotations $R_1, R_2$ such that $P[R_1 \cdot V] = P[R_2 \cdot V]$, then the action of an additional rotation $R$ can be ill-defined on $I = P[R_1 \cdot V] = P[R_2 \cdot V]$ in the sense that it becomes multi-valued. In the next sections, we explain and illustrate this issue, by showing a necessary and sufficient condition on the volume, to define a group action over the image space.

\subsection{Conditions on the volume}

We propose a necessary and sufficient condition on $V$ to define a proper 
group action of $SO(d)$ on the image space $\mathcal{I}_V$.

\begin{lemma}\label{lem:geometric_constraints}
 
Let $V$ be a volume of $L_2(\Omega^{(d)})$. Consider the map
 \begin{equation}\label{eq:rho}
      \begin{array}{cccccc}
         \rho & : & SO(d) \times \mathcal{I}_{V} & \to & \mathcal{I}_{V} \\
         & & (R , I) & \mapsto & P [(R \circ R_{I}) \cdot V ], \\
         \end{array}\\
 \end{equation}
 where $R_I$ is a rotation such that $I = P[ R_{I} \cdot V]$. Then, $\rho$ defines a group action of $SO(d)$ on $\mathcal{I}_V$ if and only if $V$ is such that for all $R_{1}  , R_{2} \in SO(d)$:
 \begin{align*}\tag{*}
  P [R_{1}\cdot V ] &= P [R_{2}\cdot V ]\\  
  \Rightarrow 
  \forall R \in SO(d), \ P [(R \circ R_{1})\cdot V ] &=P [(R \circ R_{2})\cdot V ].
 \end{align*}
\end{lemma}
Note that $R_I$ in Eq.~\eqref{eq:rho} exists by definition of $\mathcal{I}_{V}$ but is not necessarily unique.  

\emph{Proof}:
(Necessary condition) Assume \(\rho\) to be a group action, and let $R_1,R_2, R \in SO(d)$, such that $P[R_1\cdot V] = P[R_2 \cdot V] =I$. By definition of $\rho$ as a map (i.e. single-valued), $ P[(R\circ R_1) .V]=\rho(R,I) = P[(R \circ R_2).V]$. Thus, $V$ satisfies (*).\\
(Sufficient condition) 
Assume $V$ satisfies (*). We first verify that $\rho$ is well defined. If there exists $R_1, R_2 \in SO(d)$ such that $P[R_1.V] = P[R_2.V] = I \in \mathcal{I}_V$, then $\rho (R,I)$ is uniquely defined regardless of using $R_1$ or $R_2$, by definition of the condition (*). 
It remains to verify the identity and compatibility axioms: 

$\bullet$ Identity: $\rho(Id,I) = P[Id.R_I \cdot V] = P[R_I \cdot V] =I$.

$\bullet$ Compatibility: Let $R_1,R_2 \in SO(d).$
    \begin{eqnarray*}
    &\rho(R_2,\rho(R_1,I)) 
    = \rho(R_2,P[(R_1\circ R_I) \cdot V]) \\
    &\qquad\qquad= P[(R_2\circ R_1) \cdot R_I \cdot V] = \rho(R_2\circ R_1,I).  \qquad \square  
    \end{eqnarray*}
 
 In practice, it can be hard to determine if a volume satisfies $(*)$. Thus, we state a sufficient condition $(**)$, that allows us to find such volumes in practice. This will also show that the set of volumes satisfying (*) is not empty:
 
 \begin{lemma}\label{lem:injectivity}
 Consider a volume $V$ such that for all $R_{1}  , R_{2} \in SO(d)$
 \begin{equation}\label{eq:injectivity}\tag{**}
 P [R_{1}\cdot V ] = P [R_{2}\cdot V ] 
    \Rightarrow R_{1}=R_{2}. 
\end{equation}
Then, $V$ satisfies the geometric constraint $(*)$ from Lemma~\ref{lem:geometric_constraints}.
 \end{lemma}
 
 The proof is straightforward. This condition also implies that $\rho$ is injective as a function of $SO(d)$. Thus, the image space $\mathcal{I_V}$ can be reduced to 
 \begin{align*}
    \mathcal{I}_V &= \left\{ 
        I \in L_2(\Omega^{(d-1)}) | \exists ! R_I \in SO(d) , 
            I = P(R_I \cdot V) \right\}. 
  \end{align*}
  
 Note that volumes verifying (**) form a strict subset of volumes verifying (*): for instance, the hypersphere $S^d$ satisfies (*), but not (**).  Interestingly, volumes verifying (**) allow us to define a group action where the stabilizer is equal to the identity element (i.e. a \textit{faithful} action) and therefore, where the image space defined by Eq.~\eqref{eq:image_formation} forms an orbit that is homeomorphic to $SO(d)$ by virtue of the orbit-stabilizer theorem of group theory \cite{MiolanePoitevin2020CVPR}.

\section{Applications}

\subsection{Constructing compatible volumes}\label{sec:construction}

We construct ``compatible" volumes $V$, i.e. volumes which guarantee a group action on the image space $\mathcal{I}_V$, by verifying the condition (**). To find such a volume $V$ in $L^2(\mathbb{R}^d)$, we model $V$ as a sum of $n$ Dirac functions at $X_1,\dots,X_n \in \mathbb{R}^d$. As a consequence, condition (**) can be expressed in matrix form. More precisely, $(X_1,\dots,X_n)$ should verify that there is no distinct rotation matrices $R_1$, $R_2$ and permutation $\sigma$ of the symmetric group $S_n$ such that
\begin{equation}\label{eq:volume_condition_matrix}
 P R_1 \left(X_1 \ X_2 \ldots X_n \right) = P R_2 \left(X_{\sigma (1)} \ X_{\sigma(2)} \ldots X_{\sigma(n)}\right),
 \end{equation}
where $P$ is the matrix associated with the projection operator (e.g. $\begin{pmatrix} 1& 0\\ 0&0\end{pmatrix}$ in dimension 2), and  $\left(X_1 \ X_2 \ldots X_n \right)$ 
is a $d\times n$ matrix which $i$-th column takes the coordinates of. Upon parameterizing the rotations $R_1$ and $R_2$ (e.g. take $\theta_i$ ($i=1,2$) to define $R_i = \begin{pmatrix} \cos \theta_i & -\sin \theta_i\\ \sin \theta_i & \cos \theta_i \end{pmatrix}$ in dimension 2), one can formally solve the system of equations given by \eqref{eq:volume_condition_matrix} (for all permutations of $S_n$), and conclude that a volume   $V= \sum_{i=1}^n \delta_{X_i}$ satisfies (**) when no solution is found. In practice, covering all permutations is manually intractable. Thus, we resort in the next subsection to using solvers such as Mathematica, to show that volumes verify the injectivity condition (**).

\subsection{Implementing geometric VAEs}
\label{sec:VAE}

We illustrate the importance of our previous results in the context of the geometric VAEs introduced by Falorsi \emph{et al.} \cite{falorsi2018explorations}, by discussing the impact of having a volume $V$ that guarantees a group action of $SO(d)$. The VAE architecture of Falorsi \emph{et al.} contains an encoder that infers a rotation $R_I \in SO(d)$, and a variance $\sigma_I^2$ from an image $I$. The tuple $(R_I, \sigma_I^2)$ is used to sample a rotation $R$ using the reparametrization trick \cite{Kingma2014Auto-EncodingBayes} that is here adapted to the Lie group structure of $SO(d)$ \cite{falorsi2018explorations}. The sampled rotation $R$ is transformed into a matrix $T(R)$ through an irreducible representation $T$ of $SO(d)$. The decoder then combines a latent variable representing the oriented volume by matrix multiplication with $T(R)$, to reconstruct an image. The network is trained to minimize a loss function that combines two terms of reconstruction (via the binary cross entropy) and regularization (via the Kullback Leibler divergence) between the input and output images \cite{Kingma2014Auto-EncodingBayes}. For visualization purpose, we adapt this architecture -- originally introduced for $SO(3)$ \cite{falorsi2018explorations} -- to $SO(2)$ e.g. adapting the reparametrization trick, matrix representation and loss function. We also note that this architecture was not proof-tested in the context of the projection of a Lie group action, which is the goal of the experiments here. 

\subsection{Experiments: Pose inference}

\begin{figure}[h]
    \begin{center}
    \includegraphics[scale=0.28]{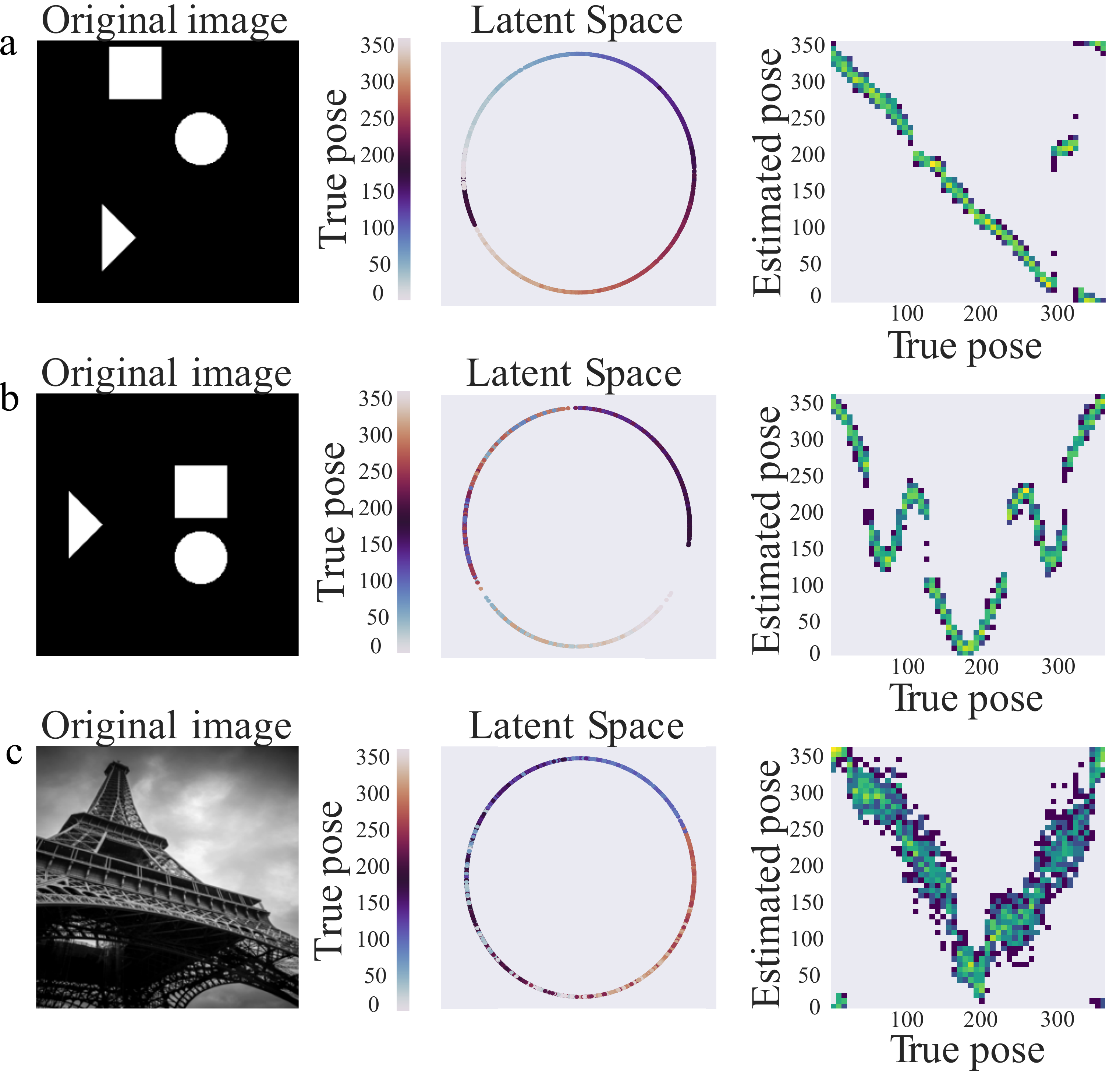}
    \end{center}
    \caption{Pose estimation in $SO(2)$. We apply the VAE described in Section~\ref{sec:VAE} to 1D lines, obtained by projecting 2D ``volumes", as original images in (\textbf{a}-\textbf{c}): (\textbf{a}) shows a ``volume" that derives from three points satisfying the injectivity condition (**); (\textbf{b}) a ``volume" similar to the one in (\textbf{a}), but that does not satisfy the condition (*); (\textbf{c}) a real picture. The middle panels show the latent space $SO(2)$, where the polar angle represents the estimated pose and the color represents the true pose. The right panels compare the true and estimated poses.} 
    \label{fig:experiments}
    \end{figure}

We considered three datasets, obtained by projection of three different 2D ``volumes'' shown in Fig.~\ref{fig:experiments}: one toy volume of three points (with simple shapes built around) that satisfies the injectivity condition (**) and thus (*) (Fig.~\ref{fig:experiments} \textbf{a}), using the construction described in Section \ref{sec:construction}, which makes it compatible with the group action; a second volume of three points similar to the first (Fig.~\ref{fig:experiments} \textbf{b}), but not satisfying (*), and a third volume from a real picture (Fig.~\ref{fig:experiments} \textbf{c}), which shows some degree of (approximate) symmetry. The theoretical development of the previous sections suggests that the pose inference to be properly performed in our first dataset, while encountering issues in the other two. We generate 2000 1D ``images" for each 2D ``volume". We train the VAE by performing a hyperparameter search on the depths of the encoder and the decoder. We study the poses inferred from the run with the best validating loss in Fig.~\ref{fig:experiments}. Comparing the estimated pose in $SO(2)$ of the VAE with the ground truth confirms that the pose inference is correctly performed in Fig.~\ref{fig:experiments} \textbf{a}, up to a reflection of the original image (which is also coherent with the definition of the projection). In contrast, the comparison between the estimated pose and the ground truth in both Fig.~\ref{fig:experiments} \textbf{b} and \textbf{c} yield a ``V-shaped" plot, which mixes up the poses at $180 \pm \theta$ degrees and distributes them between 0 and 360 degrees. As a result, the reconstructed volume will be significantly worse.

\vspace{-5mm}
\section{Conclusion and Future work}

This paper investigated the conditions under which a group action of $SO(d)$ can be defined over a space of images defined by projection of oriented volumes. We showed that such a group action is not valid in general, with a necessary and sufficient condition on the volume required to ensure it. In the context of VAEs with latent space homeomorphic to compact Lie groups, we illustrated that this condition is critical for properly inferring the pose from projected images, and thus reconstructing the original volume. As the present analysis and experiments rely on using the Radon transform in the image formation model, it could be interesting to generalize our study to any mathematical projection (as a linear operator $P$ verifying that $P \compose P = P$).

These results provide important insights on the applicability of VAEs with latent space homeomorphic to $SO(d)$. For instance, in the context of cryo-EM \cite{MiolanePoitevin2020CVPR,donnat2022deep}, biomolecules are prone to symmetries. This would preclude the group action as in our experiments, and justifies the need for more specific priors
. In principle, the Dirac functions that compose our toy models can also serve as a model for atomic structures. However, using the present approach to study if molecular volumes are compatible with condition (**) would be challenging in terms of complexity, due to the number of atoms involved and the need to cover all permutations in \eqref{eq:volume_condition_matrix}. It would also be interesting to study the effect of the image resolution, and how a loss of details at low resolution can lead the performance in pose inference to collapse, even for a volume that is theoretically compatible with the action of $SO(d)$. 

The choice of an alternative to $SO(d)$ is also interesting to study. For example, one could consider quotienting $SO(d)$ by some appropriate subgroup, with some proper linear representation and reparametrization trick. Another possible approach to address the general case where the group action cannot be defined is the following: As $SO(d)$ acts on the space of volumes one can consider the corresponding orbit $O_V$ of $V$. Denoting $G_{0, V}$ the stabilizer of $V$, the geometry of the space of oriented volumes is given by the orbit-stabilizer theorem and we have: $O_V \sim SO(d) / G_{0, V}$. For simplicity, let us assume $G_{0, V} = Id$ such that: $O_V \sim SO(d)$. By definition, the space of  images produced via the projection is $I_{V*} \sim P(SO(d))$ i.e. obtained by the projection under $P$ of the space homeomorphic to $SO(d)$. Since the projection operator $P$ is continuous, we thus have a continuous immersion of the manifold $SO(d)$  into the space $L_2(\Omega)$. Since the Hilbert space $L_2(\Omega^{(d-1)})$ is naturally equipped with a Euclidean metric, we could investigate the pullback of this metric on $SO(d)$ via this immersion $i$ and use the associated Riemannian operators to generalize the framework of Falorsi \textit{et al.} \cite{falorsi2018explorations}. As this case involves additional technical developments, we leave it for future work.

\vspace{3mm}

\textbf{Acknowledgments}: This research was supported by a Mitacs PIMS France-Canada fellowship. Computational resources and services were provided by Advanced Research Computing at the University of British Columbia.

\bibliographystyle{ieee_fullname}
\bibliography{bibli.bib}

\end{document}